# A Preliminary Exploration into an Alternative CellLineNet: An Evolutionary Approach

Akwarandu Ugo Nwachuku[1]; Xavier Lewis-Palmer[2] and Darlington Ahiale Akogo[3]

Northeastern University, Vancouver BC V6B 5A7, Canada
Old Dominion University, Norfolk, VA 23529, United States of America
minoHealth AI Labs, 00233, SCC Inside, Weija, Accra, Ghana

**Abstract:** Within this paper, the exploration of an evolutionary approach to an alternative CellLineNet: a convolutional neural network adept at the classification of epithelial breast cancer cell lines, is presented. This evolutionary algorithm introduces control variables that guide the search of architectures in the search space of inverted residual blocks, bottleneck blocks, residual blocks and a basic 2x2 convolutional block. The promise of EvoCELL is predicting what combination or arrangement of the feature extracting blocks that produce the best model architecture for a given task. Therein, the performance of how the fittest model evolved after each generation is shown. The final evolved model CellLineNet V2 classifies 5 types of epithelial breast cell lines consisting of two human cancer lines, 2 normal immortalized lines, and 1 immortalized mouse line (MDA-MB-468, MCF7, 10A, 12A and HC11). The Multiclass Cell Line Classification Convolutional Neural Network extends our earlier work on a Binary Breast Cancer Cell Line Classification model. This paper presents an on-going exploratory approach to neural network architecture design and is presented for further study.

## 1  Introduction

The source of many breakthroughs and great inventions could be attributed to an application of biological principles to the design of human systems, i.e the use of high-frequency sonars by dolphins and bats for food location and navigation applied to the development of echolocation in submarines[1]. An ability of computers to model complex systems lead to strong interdisciplinary efforts from both fields, for example, a better understanding of network properties of biological systems and connections between network structure and function can be formed, neural information could be used by machine learning classifiers to create hybrid/computer systems, flexible biological patterns such as the immune system can be modelled with computers[17].

Convolutional neural networks are a series of mathematical functions capable of pattern recognition are a principle that has been adopted from biological operations. In [18], Fukushima proposes a self-organized visual pattern recognition system that is capable of learning without a teacher, by acquiring an ability to recognize stimulus patterns based on geometrical similarity[18]. Simple and complex cells rest in layers among each other, formed in a network; together they stand as a tool capable of human-level pattern-recognition: neocognitron. The simple cell may recognize a pattern in a particular position on an image, the complex cell can recognize that same pattern in another position of an image, this ability of complex cells is termed the spatial invariance. By collecting a collection of simple cells with different positions and summing their output, complex cells can respond to any pattern anywhere on an image[19]. This notion captured the idea behind the neocognitron, by arranging a layer of simple cells followed with complex cells human-level pattern recognition was able to be captured [18].

Evolutionary algorithms first introduced by John Holland in 1970 [21], are a group of algorithms able to evolve solutions to an optimization problem. By maintaining a population of varied solutions, evolutionary algorithms determine the best solutions for a user defined fitness metric[1]. These algorithms have been successfully used in optimizing decision support in marketing strategies and long-term production planning[20], they are also presented as a solution to the travelling salesman problem in [22].

The research presented in this paper is motivated by a goal to discover a viable means to develop a neural network architecture. We explore an alternative to CellLineNet, with an evolutionary algorithmic approach in the design of a convolutional neural network fit for the classification of 5 epithelial breast cancer cell lines. We develop a new simple evolutionary algorithm: EvoCell, that searches a solution space of inverted residual blocks, residual blocks, bottleneck blocks and a simple 2x2 convolutional block for a network architecture highly accurate in the classification of breast cancer cell lines. Our approach modifies the evolutionary algorithm to weigh the importance

## 1.1 Data acquisition

Our dataset was a collection of 408 images of MDA-MB-468, MCF7, 10A, 12A and HC11 Epithelial Breast cell Lines collected from multiple publications, thereby representing multiple labs. This dataset was used in the training and validation of evolved versions of CellLineNet. The samples as seen in Figure 1 below vary within each class and yet share certain similarities across the various classes. Shape and clustering patterns give slight cues toward identification and classification, but could still pose difficulty for human eyes for work en masse, depending on confluency and other micro environmental conditions, lighting, and phase. The MCF7, MDA-MB-468, MCF-10A, MCF-12A, and HC11 cell lines are all epithelial cell lines, meaning that they line outer and inner surfaces of tissues; specifically they are mammary lines. The first four are human tissues, with the first two of those, the MCF7 and MDA-MB-468 cell lines being adenocarcinomas, cancerous cells that occur in glandular tissue. The other two, MCF-10A and MCF-12A, represent normal human epithelial tissues. The last cell line, HC11, was isolated from mice. Each is useful in a variety of studies on breast cancer.

1.2  **Data preprocessing**

As shown in Figure 1 below, the dataset contains 17 MDA-MB-468 adenocarcinomic human mammary epithelial cell images, 23 MCF7 adenocarcinomic human mammary epithelial cell images, 17 HC11 mouse mammary epithelial cell images, 21 MCF10A normal human mammary epithelial cell line images and 12 MCF12A normal mammary epithelial cell line images. To preprocess the original imageset, each were reshaped and normalized before being augmented with an addition of horizontal flips, vertical flips, random noise and random rotations. This brought our imageset to a count of, 81 HC11 cells, 84 MCF7 cells, 80 MCF10A cells, 90 MCF12 cells and 83 MDA cells, bringing the dataset to a total of 408 images to train and validate evolved CellLineNet V2 networks. Overall, a minimal data set was used to gauge performance.

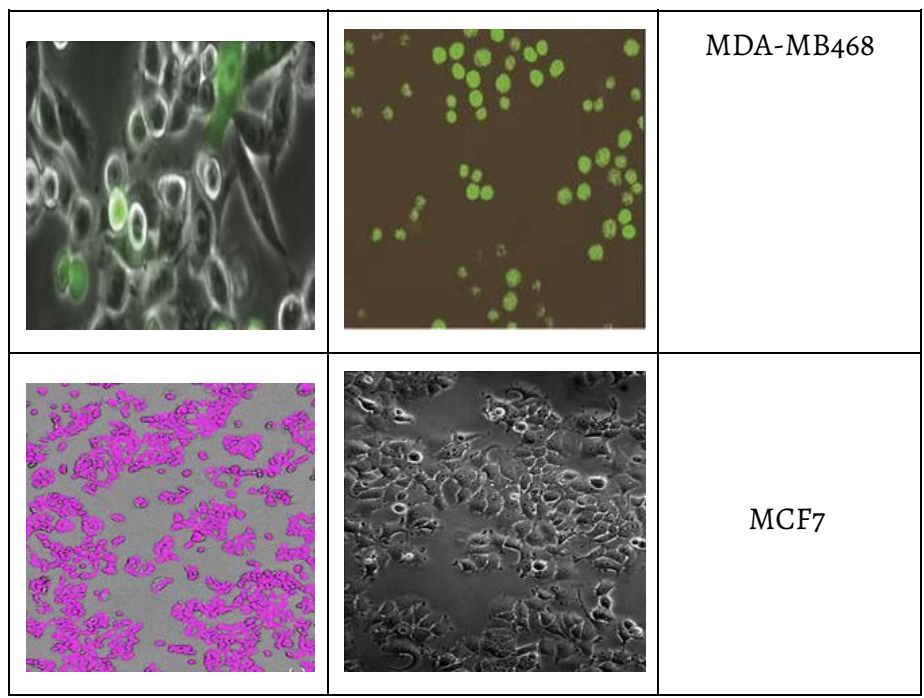

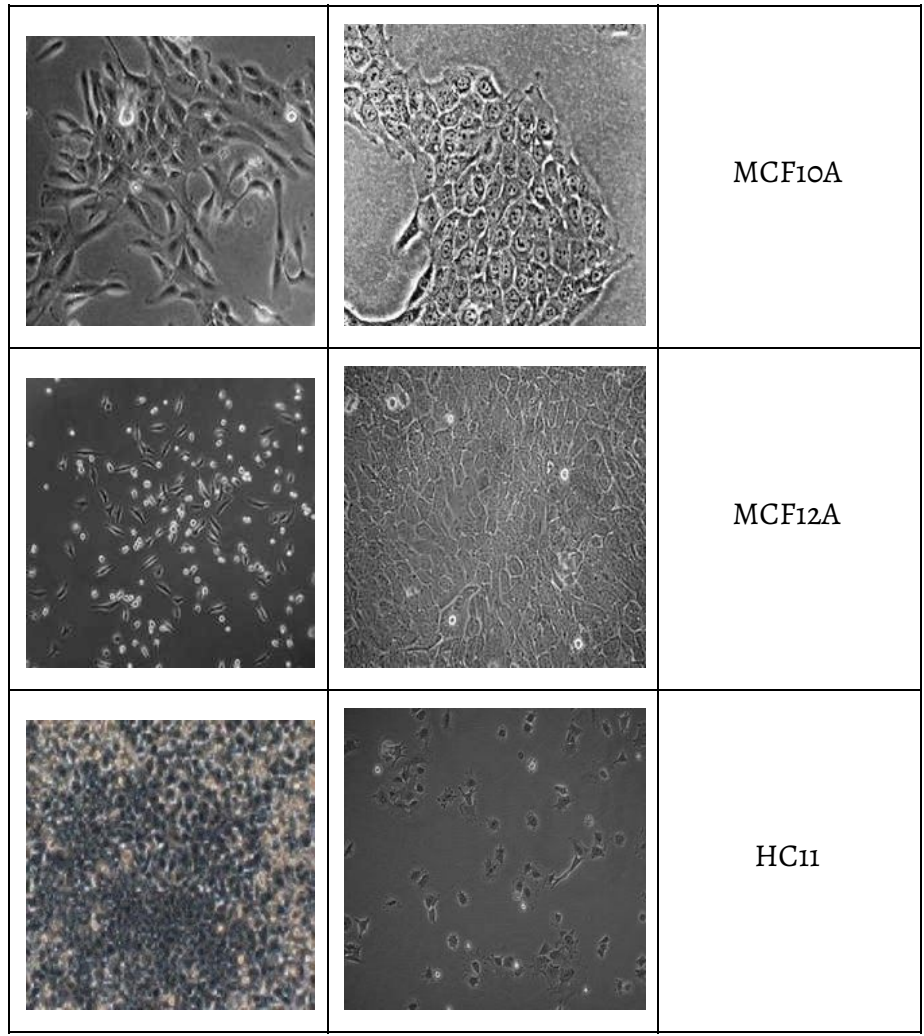

**Figure 1.** Samples of the imaged Breast Cancer Cell Lines belonging to the 5 classes (MDA-MB468, MCF7, 10A, 12A and HC11) used in training EvoCell. As shown, there exist large variations of samples within each class and some similarities across the various classes.

## 2   Literature Review

Genetic algorithms are a technique capable of solving multi-objective optimization problems. These problems are typically constrained by factors that create a search space of varied solutions explorable by an evolutionary approach. An evolutionary approach to multi-objective optimization utilizes biological concepts of mating among species and mutation in an individual species. Individuals in the mating pool are generally selected at random to mate and produce offspring and every offspring might undergo some change to become a new individual[1]. In Genetic algorithms, this process is termed crossover and mutation[1], they occur every cycle or generation, and are operations that control the direction of the search and lead the algorithm to an optimal solution. Our research employs this technique to find an optimal neural network architecture for our classification problem. Possible neural network architectures are defined in a search space with an encoding classification specific for the problem at hand; in [2] an example of a neural network individual is 32-64-0.2-64-256-0.8-512-256-256-512, where numbers between 0.5 and 1 represent pooling layers and all other numbers represent output feature maps from each layer. In [3], the authors use a factorized hierarchical search space, where each block holds the parameters of a neural network individual to be searched: convolution operation, kernel size, SERatio, skip operation, filter size, and the number of layers. In the same study, the authors find the Pareto optimal solution with the use of reinforcement learning, particularly because of the convenience it offers.

An interesting take on building feature extracting blocks are Real, Aggarwal, Huang, and Le's method in representing the blocks of architecture in the NASNet space. The NASNet space achieves 83.9% top-1/96.6% top-5 ImageNet accuracy, a new state of the art record for any automatically determined model. Their evolutionary strategy implements a method built on a graph search space, where labels represent convolution operators, vertices represent hidden states and labelled edges common network operations. Their strategy implements a random search through the solution space, it does so by randomly reconnecting edges and labels on vertices[3]. They leave two free parameters in their search algorithm, that act as a buffer between the final network and the initial stage of architecture evolution(constructs of internal operations on blocks per stage). The initial stage utilizes a graph or tree to represent the sequence of blocks in the neural network. The common principle of ageing, implemented in evolutionary algorithms, is to keep models for a longer period in the loop of evaluation while evolving the population, is applied in [3].

Other fields of Artificial Intelligence have devised an automatic approach to neural network architecture, Zoph and Le conduct neural network architecture search with reinforcement learning, they generate architecture hyper-parameters with a controller recurrent network, they train the controller with reinforcement learning--essentially finding the best neural network architecture[4]. Develop a more complex model by adding skip connections or different layer types, and finally define operations within layers with predictions from the controller RNN[4]. The same authors continue this study in hyper-parameter optimization, as they borrow much from LSTMs and Neural Network Architecture Search Cell to develop NasNet: the best performing model trained in the NASNet space[5]. Other approaches in the search for automated design of neural network architecture include differentiable search[6] and other learning algorithms[7-9].

The standard and most successful approach to architecture design is the hand-crafted approach, an approach that has yielded highly accurate and efficient models[10]. Such models include: SqueezeNet [13] operates with a low number of parameters and computation due to its use of 1x1 convolutions and a reduced filter size; MobileNet [12] extensively employs depthwise separable convolution to minimize computation density[10]; ShuffleNets [11] use pointwise group convolution and channel shuffle with very limited computing power and demonstrate superior performance over other structures by maintaining accuracy[11]. Condensenet [15] allows for efficient computation with a model designed with standard group convolutions. Recently, MobileNetV2 [14] achieved state-of-the-art results among mobile-size models by using resource-efficient inverted residuals and linear bottlenecks[10]. Due to the large design space for these models, it is a challenge that takes significant human effort to achieve the successful design of these models; our paper presents a methodology to the design of neural network model architectures, with the help of an evolutionary algorithm(EvoCell) and design space(BlockSearch), we are able to design a convolutional neural network trained in the accurate detection of 5 types of epithelial breast cell lines[16].

## 3  Overview: EvoCell

In this section, an encoding scheme is presented to represent individual genes in our population. This is followed by the presentation of pseudocode of the evolutionary algorithm before detailing it's guiding principles and levels of operations. This is then concluded by discussing limitations of the approach, followed by the introduction of methods used to improve the performance of the model.

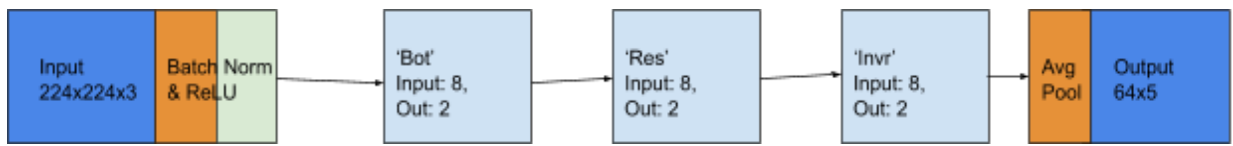

**Figure 2.** An example of the proposed encoding strategy representing a Convolutional Neural Network from our search space. The code for each layer represents the block at the layer, the input and output channels for a block in the layer.

'Bot' - Bottleneck, 'Invr' - Inverted Residuals, 'Res' - Residuals, 'CrLU'-Basic Convolution block. The layer with an 'Invr' block encodes the stride and expansion ratio for the block as well.

**Algorithm Overview**
**Algorithm 1:** The Proposed Algorithm
**Input:** Building blocks of model, population size, maximum number of generation, and image dataset

**Output:** Best performing model after max generation
$P_0$ ← Initialize a population of models built from random spread of building blocks
Gen = 0
**While** *gen < max_generation do*
    *Evaluate fitness of initial population;*
    *Update control variables if gen > 1;*
    $C_0$ ← *Generate offspring with crossover and mutation operators;*
    $P_{t+1}$ ← *Select next generation from offspring and present population;*
    $T$ ← $t + 1$
**end**
**Return** best performing individual

---

EvoCell: an evolutionary approach to Neural Network architecture, presents a genetic algorithm optimized to find accurate convolutional neural network models in a search-space, the algorithm is guided by two principles established: block presence ratio and set torques. The first principle builds confidence toward the presence of certain blocks within the architecture, whereas the second principle builds confidence toward certain combinations of blocks in the architecture. The formula below details the calculation of the block presence ratio for a bottleneck block and the block presence ratio for a residual block from any model in the search space;

$$Bottleneck_{presence\_ratio} = \frac{\# \text{ of bottleneck blocks}}{\# \text{ of blocks in model}} * model\_accuracy$$

$$Residual_{presence\_ratio} = \frac{\# \text{ of residual blocks}}{\# \text{ of blocks in model}} * model\_accuracy$$

Since only the arrangement of 4 blocks is optimized, one can easily keep track of the 16 possible combinations of blocks in sets(where sets are a makeup of two blocks). This leads to the initialization of a control variable: torque. The torque in a model exists for every set in the model, it is a means to measure the combination strength between sets of blocks in the model. Control variables: Block presence ratio and torque, are calculated after training and evaluating the performance of a model in the search space. After the initial generation has developed offspring and mutated an evaluation is run on all individuals, the torque for a set in the model is calculated as follows;

$$T_{BR} = \# of\ sets\ in\ model * model\_accuracy$$

Where BR stands for bottleneck block and Residual block respectively. These control parameters aid the crossover and mutation search operators.

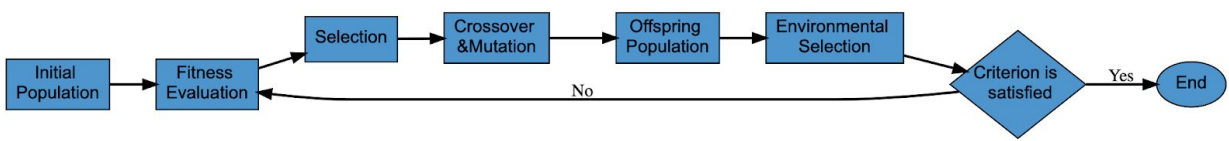

**Figure 3.** Flowchart of genetic algorithm [2]

In the crossover operator, when similar sets of blocks are present in both parents, the highest torques within those similar sets are used as a heuristic for the crossover operator to determine what position to split the parent genes. Since both parents have the same set, finding the highest torque and splitting at that position becomes possible. Crossover only happens 40% of the time and the second stage of crossing parents at the position of highest torque happens 50% of the time, the other 50% is a cut at a random location for any parent. The mutation operator occurs 60% of the time in a generation and any of the four options may occur: addition of the block with the highest presence ratio at a random location in the model, addition of a random block from our selection of blocks, removal of a random block in the model and removal of the block with the lowest presence ratio in the model.

---

**Algorithm 2:** Crossover operation
**Input:** Two parent individuals from population
**Output:** Two offspring generated
*set_1 , set_2 ← extract block sets from parents genes*
$Prob_{similar\_blocks\_crossover}$ ← *Random number generated* (1,10)
**If** $Prob_{similar\_blocks\_crossover} > 5$
    *Find similar sets between both blocks*
    *If no_similar_sets == False :*
        *Find highest torque in similar sets*
        *Crossover at highest torque position*
**If** $Prob_{similar\_blocks\_crossover} < 5$ *or no_similar_sets == True :*
    *Select a random parent to be front half of gene*
    *Select a random parent to be second half of gene*
    *Crossover at random position in gene*
**End**
**Return** two offsprings

---

**Algorithm 3:** Mutation operation
**Input:** One individuals from population
**Output:** One mutated individual
*[1, 2, 3, 4] ← Select random choice from list*
**If** *Choice == 1 :*
    Add block with the highest presence ratio
**If** *Choice == 2 :*
    Remove block with the highest presence ratio

**If** *Choice* == 3 :
    Add random block from selection
**If** *Choice* == 4 :
    Remove random block
**End**
**Return** mutated individual

---

After each generation, EvoCell eradicates the lowest-performing models, by assessing their validation accuracy scores. The actual solution EvoCell searches for is the proper architecture of a network that consists of these blocks: Inverted residual blocks, residual blocks, bottleneck blocks, and 2x2 convolutional blocks. What EvoCell fails to do is optimize on the feature maps between blocks in the network. We believe this is a source of the 53.4% accuracy model generated after 15 generations of evolution. The inability to extract more features per block, due to resource constraints, holds back achieving higher accuracy measures.

## 4  Results

After ~15 generations, a saturation in accuracy across all models is noticed, in addition to this, an increase in training time is noticed as well. Furthermore, the inability to automatically modify the input and output channels within the network, suggests our accuracy may only be improved by modifying input and output channels for each layer in our network. So, after 15 generations a new population was selected from the top-performing models in all generations, followed by an evaluation of their fitness after initializing specific input and output channels per layer and training for another 10 generations. A saturation in model performance was observed, after the first 4 generations; nonetheless, the accuracy of the models improved and a 50% average accuracy is maintained across all models per generation. The graph below presents the top-performing model after 25 generations of evolution in the solution space.

| Input | Operator | Output | Stride | Expand Ratio |
|---|---|---|---|---|
| $224^2 * 3$ | Conv2d | 16 | - | - |
| 16 | Invr | 24 | 1 | 1 |
| 24 | Invr | 32 | 1 | 1 |
| 32 | Invr | 64 | 1 | 1 |
| 64 | Invr | 96 | 1 | 6 |
| 96 | Invr | 160 | 1 | 6 |
| 160 | Invr | 320 | 1 | 6 |
| 320 | Invr | 1280 | 1 | 6 |
| 1280 | Avg pool 8x8 | - | - | - |
| 1280 | Linear | 640 | - | - |
| 640 * 64 | Linear | 64 | - | - |
| 64 * 5 | Linear | 5 | - | - |

**Table 1.** Outline of the evolved model after 25 generations. EvoCell's best performing model achieved 60% validation accuracy on the dataset; it consists of 7 inverted residual blocks, 1 pooling layer and three dense/linear layers.

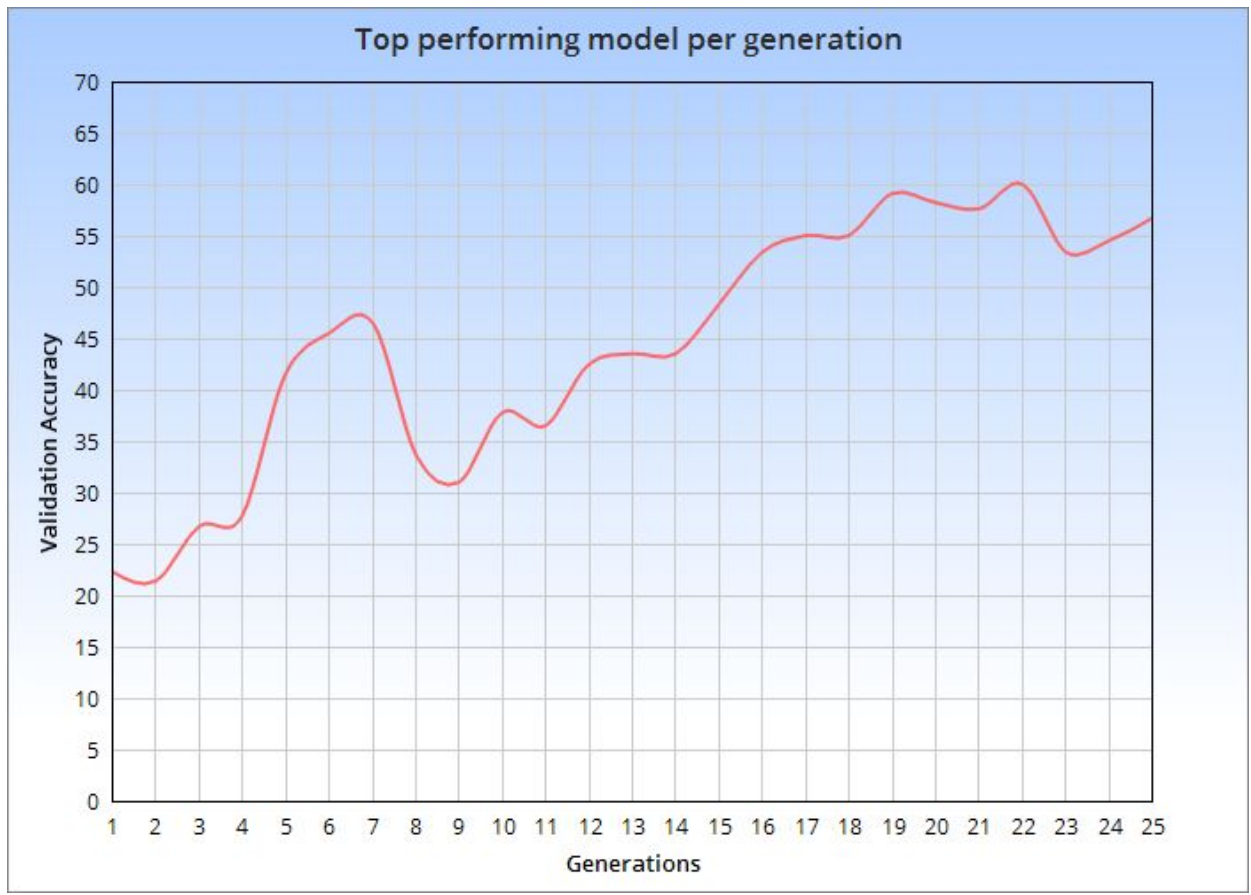

**Figure 4.** Graphical representation of the best performing models in each generation of the EvoCell evolution process.

After hand-picking models for reproduction in EvoCell, the best performing model found achieved a 60% validation accuracy on the dataset; it consists of 7 inverted residual blocks, 1 pooling layer and three dense/linear layers. In comparison to other genetically optimized neural networks, the genetically designed convolutional neural network in [2] achieved a 77.97% accuracy on the CIFAR100 dataset. The genetic algorithmic approach boosted accuracy on the dataset in [2] as the manually developed network performed at a 66.29% accuracy. In [4] a neural network architecture was determined via a reinforcement learning method, 12,800 architectures were trained and validated before a small grid search was run on certain parameters. The resulting network (with max pooling and filters) achieved an error rate of 3.65%.

## 4.1 Discussion

We believe that achieving a more accurate model within the EvoCell search space, requires updating the evolutionary process to optimize the input and output features per block in a model and more resources (i.e RAM) to train models on. The predecessor to this model trained for the specific task of epithelial breast cancer cell line detection, CellLineNet v1, was built on top of MobileNets v1 and achieved a very high accuracy of 97% [16]. One reasonable route to follow would be to build CellLineNets v2 on top of MobileNets v2, however, it was preferred to explore the possibility to design a new architecture with an evolutionary approach that was useful with variation in imaging modalities. This exploration resulted in a model that achieved 60% accuracy in the detection of epithelial cell lines on a much-reduced image load, taken from a wide selection of papers versus raw images from one or a few labs. The EvoCell evolutionary process is a novel means of exploration as the defined exploration operators- crossover and mutation, present a reward system for high performing blocks and combinations of blocks, and also a methodology to determine block performance strength in a given classification task.

We believe our explorative study has a theoretical impact in the initial design of neural network architectures for specific use cases, we also believe the search space of EvoCell could be expanded to include a wider variety of convolutional block architectures and inclusion of input and output channels in the optimization task. Furthermore, we believe researchers taking this approach of evolving network architectures would have an initial idea of convolutional blocks involved in their proposed model but not a direct solution to the depth of their network, or a specific layer of architectures. Finally, the control variables we introduce highlight the potential to direct the search of a neural network design via an evolutionary search algorithm. We recognize the model developed does not achieve very high accuracies, thus the search for a comparable, alternative approach to CellLineNet carries on, leaving us room to improve on the searching mechanisms of our evolutionary algorithm: EvoCell.

EvoCell represents another meaningful step in another direction as well: working with images compiled from more than just a few sources or imaging styles. Images being from a wider set of labs introduces more variation into picture quality, affecting lighting, degree of focus, magnification, and more, due to biases of individual lab aesthetics. This can create potential difficulties in recognition of cell lines in images by human eyes without labeling, even in narrow publications looking at a few lines, affecting assessment during peer review. A deeper importance within this exists in that this algorithm suggests the potential to obtain recognition of a cell line used regardless of the lab taking the original images and in improving said recognition via an evolutionary architecture, economically. In other words, In terms of use, It's possible that such a tool can aid means of finding publication errors in which the wrong cell type may have been posted for a non-assayed figure, but missed critical review, especially in lower quality journals. Also, this tool may be adapted in identifying lab collaboration through aligning for bias in figure aesthetics similar to a reverse image search. However, such remains to be seen as this is a preliminary work and is still in progress. Larger and more diverse training sets are expected to be helpful towards further work.

## 5  Conclusion

The research presents an exploration into the design of a neural network architecture via an evolutionary algorithmic approach. The network is constructed from four main building blocks of inverted residual blocks, residual blocks, bottleneck blocks and a simple 2x2 convolutional block, with an optimization goal to find the depth of layers in a network and the arrangement of blocks per layer in a network. The research incorporates a control strategy that monitors the performance of blocks in a network in order to increase or decrease the presence of a block. The evolutionary strategy: EvoCELL, trained and tested different neural network architectures on a 25 gigabyte RAM notebook and achieved a network capable of classifying breast cancer cell lines at 60% accuracy. The current state of the research is that the limited amount of resources available are not enough to optimize networks with more features extracted on deeper networks. In addition to this, the optimized networks were trained on a dataset encompassing images of cells taken from different labs under different conditions, as extracted from multiple journal articles [24-39].

The introduction of control variables in the optimization search process enlightens the reader on a methodology for guiding an evolutionary search. EvoCell also presents a new specification for representing neural networks as individuals in an evolutionary scheme, it is a technique applicable on a selection of feature extracting blocks and in a variety of classification task domains. The primary motivation to discover a neural network design approach via evolutionary search has proven applicable to this specific task, future directions in the research are to bypass the initial limitation from the dataset, and apply the strategy to the classification of blood cell types or others, considering a new set of feature extracting blocks [23].